\pdfoutput=1

\documentclass[11pt]{article}

\usepackage[final]{acl}

\usepackage{times}
\usepackage{latexsym}

\usepackage[T1]{fontenc}

\usepackage[utf8]{inputenc}

\usepackage{microtype}

\usepackage{inconsolata}

\usepackage{amsmath}
\usepackage{amssymb}
\usepackage{amsfonts}
\usepackage{amsthm}
\usepackage{mathrsfs}
\usepackage{mathabx}
\usepackage{bm}

\usepackage{booktabs}
\usepackage{multirow}
\usepackage{tabularx}
\usepackage{arydshln}
\usepackage{makecell}
\usepackage{colortbl}
\usepackage{siunitx}

\usepackage{graphicx}
\usepackage{svg}
\usepackage{caption}
\usepackage{subcaption}
\usepackage{wrapfig}
\usepackage{floatflt}
\usepackage{algorithm}
\usepackage{algorithmicx}
\usepackage{algpseudocode}

\usepackage{microtype}
\usepackage{nicefrac}
\usepackage{enumitem}
\usepackage{url}
\usepackage{xurl}
\usepackage{authblk}
\usepackage{lipsum}

\usepackage{soul}

\usepackage[most]{tcolorbox}

\usepackage{fontawesome5}
\usepackage{CJKutf8}
\usepackage{hyperref}
\usepackage{bbding}
\usepackage{pifont}
\usepackage{wasysym}
\usepackage{tabularx}

\title{R1-RE: Cross-Domain Relation Extraction with RLVR}
\author{
    Runpeng Dai$^{1}$\quad
    Tong Zheng$^{2}$ \quad
    Run Yang$^{3}$ \quad
    Kaixian Yu$^{4}$ \quad
    Hongtu Zhu$^{1\dag}$ \\
    $^{1}$University of North Carolina at Chapel Hill \quad
    $^{2}$University of Maryland, College Park \\
    $^{3}$BiliBili \quad
    $^{4}$Insilicom LLC \\

    \texttt{\{runpeng, htzhu\}@email.unc.edu} \\
    \texttt{tzheng24@umd.edu} \\
    \texttt{yangrun@bilibili.com} \\
    \texttt{kaixian@insilicom.com}
}

\begin{document}
\maketitle

\begin{abstract}
Relation extraction (RE) is a core task in natural language processing. Traditional approaches typically frame RE as a supervised learning problem, directly mapping context to labels—an approach that often suffers from poor out-of-domain (OOD) generalization. Inspired by the workflow of human annotators, we reframe RE as a reasoning task guided by annotation guidelines and introduce \textbf{R1-RE}, the first reinforcement learning with verifiable reward (RLVR) framework for RE tasks. Our method elicits the reasoning abilities of small language models for annotation tasks, resulting in significantly improved OOD robustness. We evaluate our approach on the public Sem-2010 dataset and a private MDKG dataset. The R1-RE-7B model attains an average OOD accuracy of approximately 70\%, on par with leading proprietary models such as GPT-4o. Additionally, our comprehensive analysis provides novel insights into the training dynamics and emergent reasoning behaviors of the RLVR paradigm for RE.
\end{abstract}

\section{Introduction}
Relation extraction (RE)~\citep{Xie2021RevisitingTN, zhao2024comprehensive} is a fundamental task in natural language processing (NLP) that involves either classifying the relations between pairs of entities (relation classification) or extracting (subject, relation, object) triples (triplet extraction) from context. RE serves as a foundation for numerous downstream applications \citep{nayak2021deep}, most notably in the construction of knowledge graphs (KGs) \citep{zhong2023comprehensive}, which have demonstrated broad utility across diverse domains such as biomedical research \citep{yang2024alzheimer} and e-commerce \citep{li2020alimekg}. Accordingly, improving the reliability and accuracy of RE techniques is crucial.

\begin{figure}[t!]
    \centering
    \includegraphics[width=0.9\linewidth]{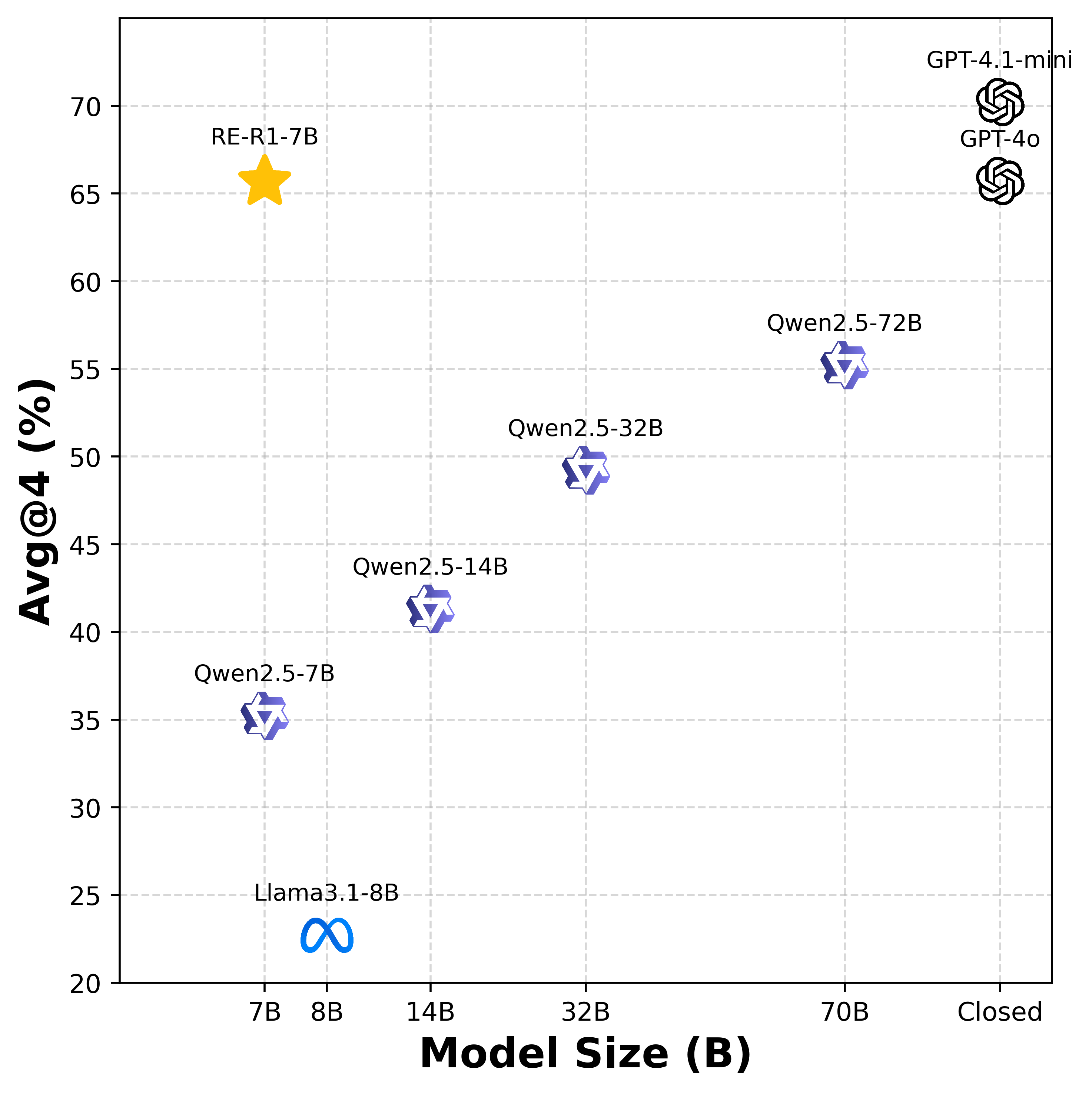}
    \caption{Testing accuracy on the MDKG dataset for R1-RE-7B trained on the Sem-2010 dataset, compared with other models. Detailed results are provided in Table~\ref{tab:RCmain}.}
    \label{fig:compare}
\end{figure}

Standard approaches to RE typically adopt a pretraining-and-finetuning paradigm, employing mid-sized pre-trained models such as BART~\citep{lewis2019bart} and BERT~\citep{devlin2019bert}. While these models are lightweight, they require substantial fine-tuning to generalize to out-of-domain scenarios. Recently, there has been growing interest in leveraging LLMs for RE, capitalizing on their rich language understanding~\citep{zeng2023evaluating} and impressive in-context learning capabilities~\citep{brown2020language}. In this context, some studies have explored few-shot prompting strategies with API models\footnote{API models refer to proprietary, closed-source models that are typically accessible only via external APIs.}~\citep{wadhwa2023revisiting, xu2023unleash}, while others have investigated supervised fine-tuning (SFT) methods~~\citep{ettaleb2025contribution} using lighter-weight, open-source LLMs for RE tasks.

\begin{figure}[htbp]
    \centering
    \begin{subfigure}[b]{0.48\linewidth}
        \centering
        \includegraphics[width=\linewidth]{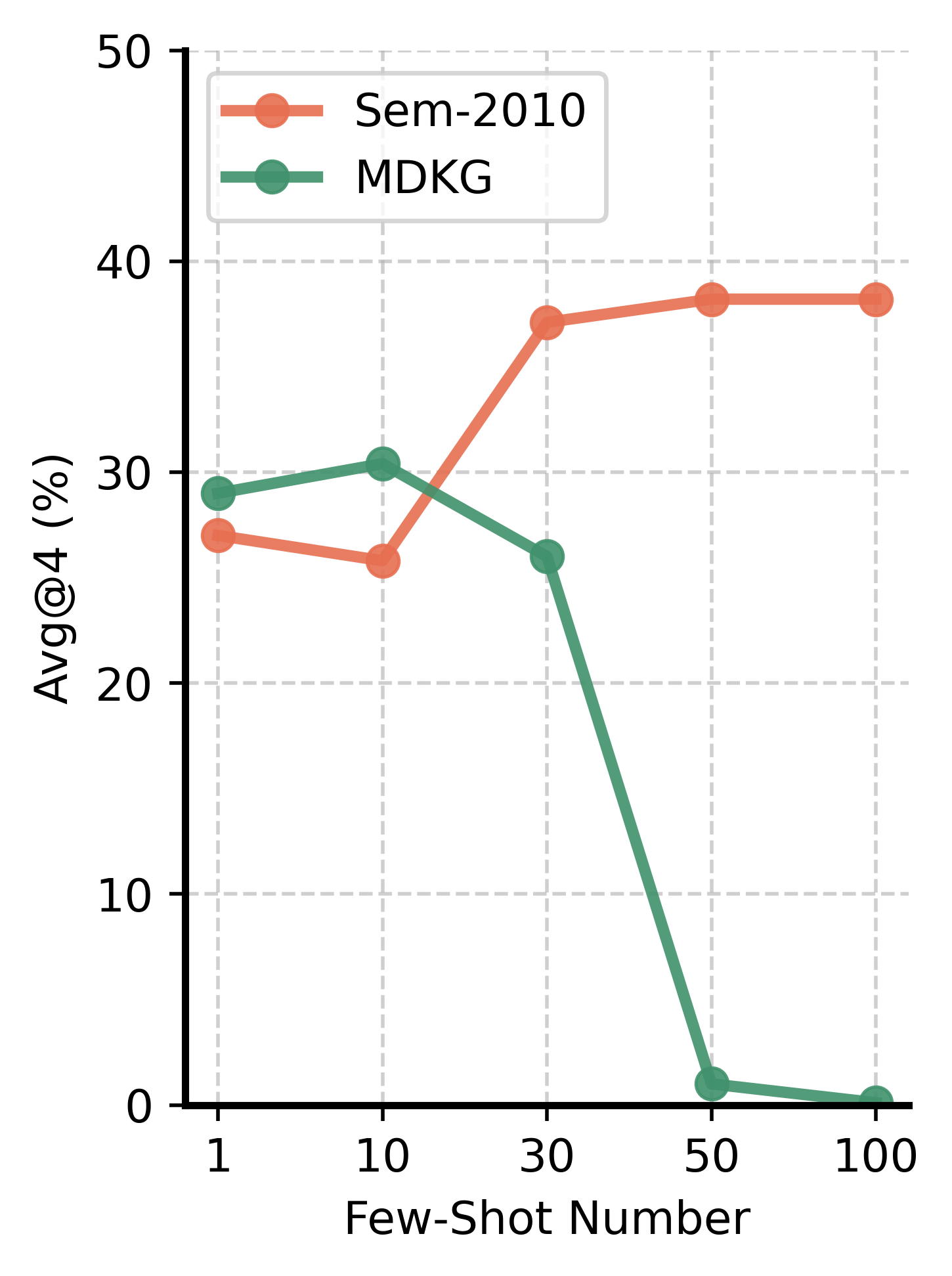}
    \end{subfigure}
    \hfill
    \begin{subfigure}[b]{0.48\linewidth}
        \centering
        \includegraphics[width=\linewidth]{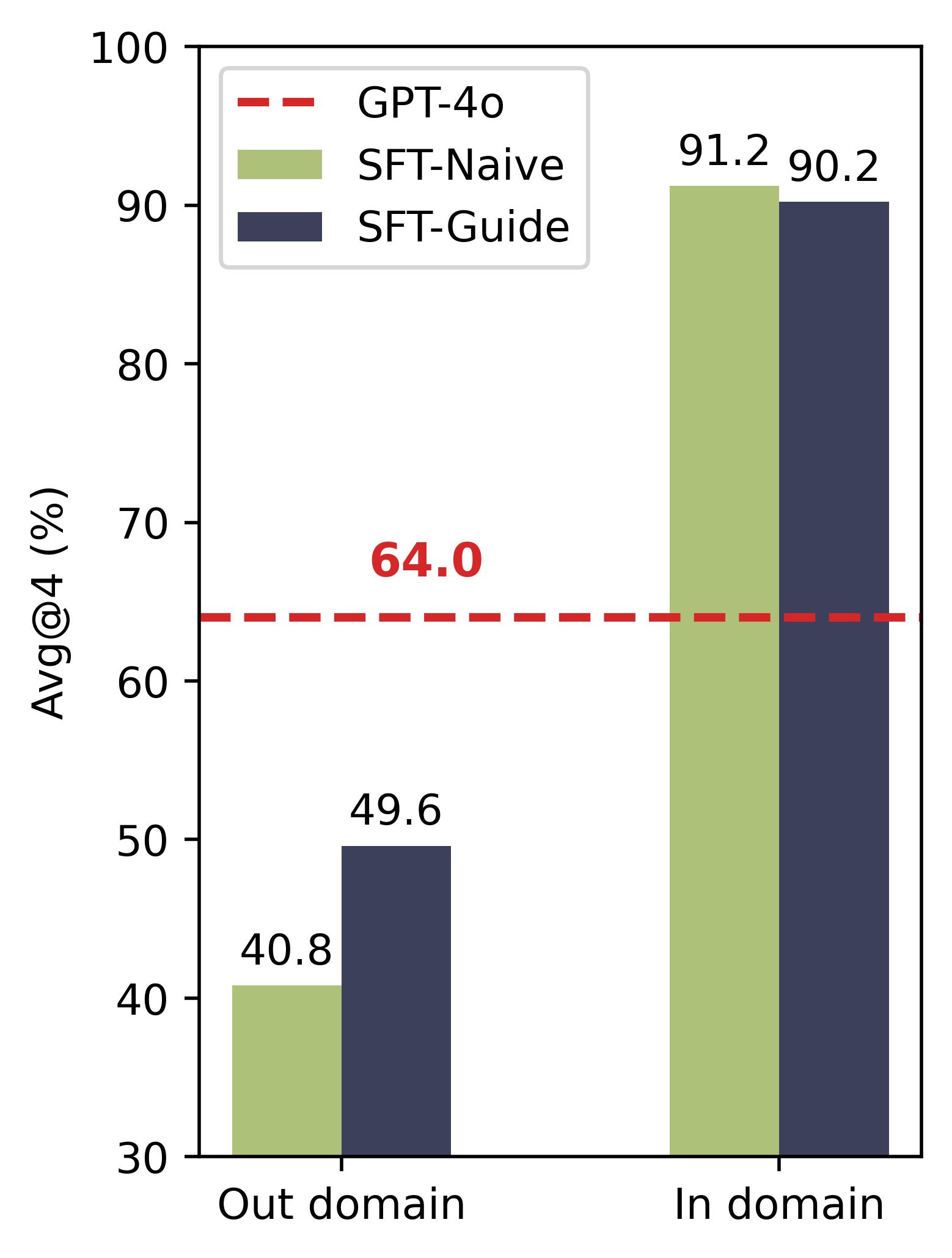}
    \end{subfigure}
    \caption{(a) Test accuracy on two RE datasets with varying numbers of few-shot examples. (b) In-domain and out-of-domain test accuracy of the SFT model using a naive prompt (Figure~\ref{fig:human}) versus a prompt incorporating the annotation guide (Figure~\ref{box:RC_template}).}
    \label{fig:two_plots}
\end{figure}

As illustrated in Figure~\ref{fig:two_plots}(a), few-shot learning yields only marginal improvements over the original model. Furthermore, Figure~\ref{fig:two_plots}(b) demonstrates that the naive SFT approach can enhance in-domain performance but provides limited generalization to out-of-domain scenarios, suggesting that SFT primarily promotes memorization rather than genuine annotation ability~\citep{chu2025sft}. This observation raises a critical question: \textit{How to enhance the RE capability of small LLMs for robust cross-domain performance?}

This observation raises a critical question: \textit{How to enhance the RE capability of small LLMs for robust cross-domain performance?} To this end, we examined the workflow of human annotators (see Figure~\ref{fig:human}). Annotators are typically provided with detailed annotation guidelines and iteratively compare the target sentence against these guidelines—formulating hypotheses, verifying them, and ultimately reaching a conclusion. While entities and relation types may vary across domains, the reasoning skills developed through annotation guideline usage are broadly generalizable.


Human intelligence motivates us to conceptualize RE as a reasoning process anchored in annotation guidelines. However, eliciting such reasoning skills from small language models is challenging. As a first step, we adapt the standard SFT paradigm by explicitly incorporating the annotation guide into the prompt. As shown in Figure~\ref{fig:two_plots}, this approach yields roughly a 10\% boost in out-of-domain RE performance; nevertheless, a substantial gap remains relative to proprietary API models. To address this, we introduce \textbf{R1-RE}, a novel framework designed to further enhance the reasoning capabilities of LLMs for RE tasks. Our approach is inspired by recent advances in reinforcement learning with verifiable reward (RLVR), which has demonstrated strong potential for promoting reasoning in smaller models on complex domains such as mathematics and code generation~\citep{guo2025deepseek}.

We evaluate our proposed method on both a public dataset (SemEval-2010 Task 8) and a private dataset (MDKG). Our results show that R1-RE improves the OOD performance of Qwen2.5-7B-Instruct by up to \textbf{+30 pp}. Notably, our 7B model achieves performance comparable to GPT-4o on MDKG. Further analysis show that: (1) R1-RE elicits genuinely human-like annotation behavior. (2) Incorporating additional training data can lead to even greater performance gains. (3) The training process of R1-RE preserves the model's performance on other tasks.


\section{Preliminaries}
\subsection{Task Definition}
Let $\mathcal{S}$ denote the input sentence, and let $\mathcal{E}$ and $\mathcal{R}$ represent the predefined sets of entity types and relation types, respectively. Let $\mathcal{K}$ denote the annotation guideline for the task, which provides the definitions of $\mathcal{E}$ and $\mathcal{R}$ as well as other annotation instructions. We consider two relation extraction (RE) tasks: Relation Classification (RC) and Triplet Extraction (TE). In this work, we primarily focus on \textbf{RC} tasks; leaving the discussions on  \textbf{TE} task in Appendix~\ref{appendix:TE}.

\textbf{Relation Classification(RC)}:  
In this task, the subject-object pair ($e_{\text{sub}}$, $e_{\text{obj}}$) has already been identified from $\mathcal{S}$. The goal is to assign a relation type $y \in \mathcal{Y}$, where $\mathcal{Y}$ is the predefined set of relation types, based on the sentence $\mathcal{S}$ and the annotation guideline $\mathcal{K}$. Each sentence $\mathcal{S}$ contains exactly one entity pair. For instance,

\begin{tcolorbox}[colback=white, colframe=black!20!white, boxrule=0.4pt, arc=1mm, left=1mm, right=1mm, top=0.5mm, bottom=0.5mm]
\textbf{\texttt{Sentence}} <e1>Counseling interventions</e1> can be effective in preventing <e2>perinatal depression</e2>.\\
\textbf{\texttt{Gold standard}} treatment-for(e1, e2)
\end{tcolorbox}

\subsection{Group Relative Policy Optimization (GRPO)}
\label{sec:GRPO}
In this work, we formulate the language generation process of LLMs as a sequential decision process and optimize the LLM with the Group Relative Policy Optimization (GRPO) algorithm~\citep{shao2024deepseekmath}.

Specifically, let $\pi_{\theta}$ denote the LLM with parameters $\theta$. At each training step, given a prompt $q$ sampled from the dataset $\mathcal{D}$, we use $\pi_{\theta}$ to generate a group of $G$ candidate outputs, denoted as ${o_1, o_2, \ldots, o_G}$. For each candidate output, we compute the corresponding reward $r_1, r_2, \ldots, r_G$ by comparing the output with the gold standard. The advantage at the $t$-th token of the $i$-th output is then calculated as
\begin{equation*}
A_{i, t} = A_{i} = \frac{r_i - \operatorname{mean}(r_1, r_2, \ldots, r_G)}{\operatorname{std}(r_1, r_2, \ldots, r_G)}.
\end{equation*}
Let $\pi_{\theta_{\mathrm{old}}}$ denote the model from the previous training step, and $\pi_{\mathrm{ref}}$ denote the original model prior to training. GRPO maximizes the following objective function to optimize $\pi_{\theta}$:  
\begin{equation*}
\begin{aligned}
&\mathbb{E}_{q \sim \mathcal{D}, \{o_i\}_{i=1}^G \sim \pi_{\theta_{\mathrm{old}}}} \biggl[
  \frac{1}{G} \sum_{i=1}^G \frac{1}{|o_i|} \sum_{t=1}^{|o_i|} \\
& \min\Bigl(
    \frac{\pi^{i,t}_{\theta}}{\pi^{i,t}_{\theta_{\mathrm{old}}}}\,A_i,\, 
    \mathrm{clip}(
      \frac{\pi^{i,t}_{\theta}}{\pi^{i,t}_{\theta_{\mathrm{old}}}},
      1-\varepsilon,\,
      1+\varepsilon) A_i \Bigr)  \\
  &-\,\beta\,D_{\mathrm{KL}}\bigl(\pi_{\theta}\,\big\|\,\pi_{\mathrm{ref}}\bigr)
\biggr],
\end{aligned}
\label{eq1}
\end{equation*}
where $\pi^{i,t}_{\theta} = \pi_{\theta}(o_{i,t} \mid q, o_{i,1} \ldots, o_{i,t-1})$, and similarly for $\pi^{i,t}_{\theta{\mathrm{old}}}$. The hyperparameters $\varepsilon$ and $\beta$ control the ratio clipping threshold and the weight of the Kullback–Leibler (KL) divergence penalty, respectively. Specifically, ratio clipping mitigates the risk of large, destabilizing policy updates, while the KL penalty constrains the updated policy from deviating excessively from the reference model $\pi_{\text{ref}}$. For additional details on the formulation of such sequential decision process and the GRPO algorithm, please refer to~\citep{yuan2025s, yu2025dapo, yue2025vapo}.

\section{Method}
In this section, we first introduce a new paradigm for relation extraction inspired by the human annotation process (Section~\ref{sec:REparadigm}). We then present the R1-RE framework (Section~\ref{sec:R1RE}) along with its associated reward design (Section~\ref{sec:reward}).

\subsection{Human-Inspired RE paradigm}
\label{sec:REparadigm}
\begin{figure*}[h]
    \centering
    \includegraphics[width=\linewidth]{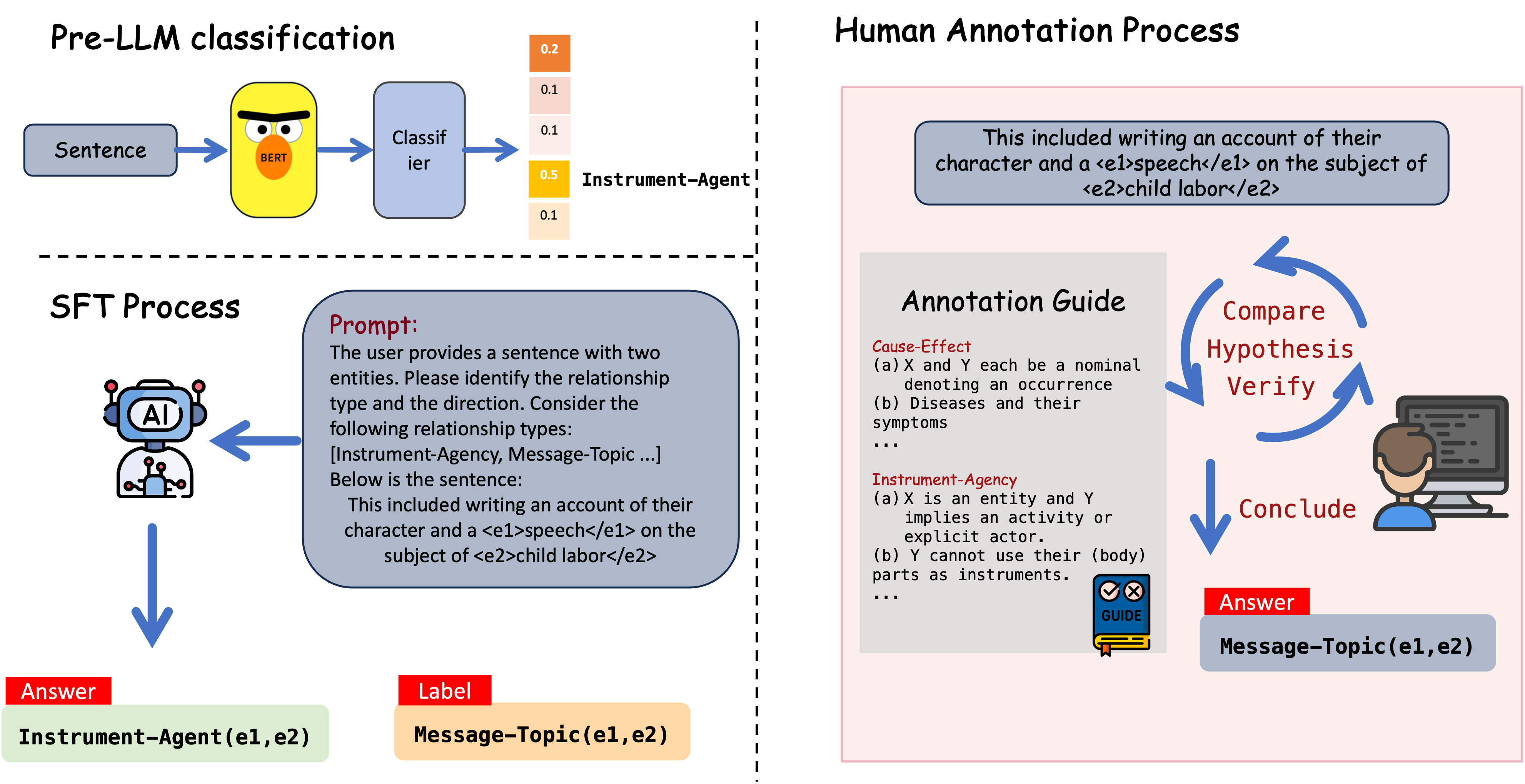}
    \caption{A comparison of existing RE training paradigm and the annotation process of human annotators.}
    \label{fig:human}
\end{figure*}

As illustrated in Figure \ref{fig:human}, existing relation extraction methods always focus on learning mappings between sentences and golden labels, for example, golden relations or gold standard annotations. Despite achieving strong in-domain performance, these direct mapping strategies always suffer from poor generalization to out-of-domain scenarios.

In contrast, human annotation is inherently a multi-step reasoning process. As shown in Figure~\ref{fig:human} (right), annotators consult an annotation guide—a manual that precisely defines each relation and entity type and provides detailed instructions for edge cases (see Table \ref{table:MDKG-pair}). During annotation, they repeatedly compare the target sentence with these definitions, formulate hypotheses, and iteratively refine their judgments through several inference steps before reaching a final decision. While concrete definitions may vary across tasks, this step-by-step reasoning paradigm is universal, making it widely applicable to diverse RE settings. Motivated by this insight, we introduce a RE framework that explicitly embeds the step-by-step human annotation process into LLM-based relation extraction, thereby bridging the gap between human reasoning and automated learning. Specifically, we introduce the following prompt, designed to guide LLMs toward human-style reasoning in relation extraction.

\begin{tcolorbox}[
    colframe=pink!80!black,
    colback=pink!10!white,
    fontupper=\small,
    boxrule=0.4mm,
    left=2mm, right=2mm, top=0.5mm, bottom=0.5mm,
    arc=2mm,
    enhanced
]
The user provides a sentence containing two entities: one enclosed in <e1> </e1> tags and the other in <e2> </e2> tags. Please identify the relation type and determine the direction of the relation between these two entities. Below are the definitions and restrictions for all relation types:\\
\{{\color{red}Annotation guide}\}\\
Please thinks about the reasoning process in the mind and then provides the user with the final answer. The reasoning process and final class are enclosed within <think> </think> and <answer> </answer> tags, respectively. The answer should aligh with format: For example, <answer> Product-Producer(e1,e2) </answer> means e1 is a product of e2 while <answer> Product-Producer(e2,e1) </answer> means e2 is a product of e1. Always use "e1" and "e2" in place of the actual entity names. Below is the sentence:\\
\{{\color{red}Sentence}\} 
\end{tcolorbox}

\noindent\begin{minipage}{\linewidth}
\captionof{figure}{The prompt for RC tasks.}
\label{box:RC_template}
\end{minipage}



\subsection{R1-RE}
\label{sec:R1RE}
Our goal is to elicit human-like annotation behaviour in the LLM-based RE pipeline. Achieving this, however, is far from trivial. Preliminary experiments show that simply prompting an LLM to emulate the annotation procedure yields unsatisfactory results. As illustrated in Figure \ref{fig:COT}, the model drifts toward a shallow, linear chain of thought instead of the richer, multi-step reasoning observed in human annotators.

To this end, we propose R1-RE, a R1-style reinforcement learning training framework that aligns an LLM’s reasoning with the multi-step workflow of human annotators. This is motivated by the recent evidence that reinforcement learning with verifiable reward as discussed in Section \ref{sec:related}, can successfully equip LLM with the capability of performing human-like reasoning on a wide range of tasks including math\citep{shao2024deepseekmath}, coding and QA\citep{lai2025med}.



Given an LLM policy $\pi$, our goal is to maximize the expected reward:

\begin{equation}
\mathbb{E}_{(q, y) \sim \mathcal{D}, \widehat{y} \sim \pi(\cdot \mid q)}\left[ r(\widehat{y}, y) \right]
\end{equation}

where $\mathcal{D}$ is the dataset of prompt--gold label pairs, $q$ denotes a prompt, $y$ the corresponding gold-standard RE result, and $\widehat{y}$ the model-generated output conditioned on $q$. The reward function $r(\widehat{y}, y)$ measures the quality of the generated answer relative to the gold label. We optimize $\pi$ with respect to this objective using the GRPO algorithm \citep{shao2024deepseekmath}, as described in Subsection~\ref{sec:GRPO}.

\subsection{Multi-stage Reward design}
\label{sec:reward}
Reward design is a critical aspect of reinforcement learning, as it encodes feedback from the environment to guide the learning process. In this work, we adopt a rule-based reward scheme similar to those used in DeepSeek-R1-Zero~\citep{guo2025deepseek} and Logic R1~\citep{xie2025logic}. Our reward consists of two components. The first is a format reward, which checks whether the model's response adheres to a specified structure. Since the primary purpose of this component is to facilitate evaluation, we enforce only minimal requirements on the response format. The second component is the accuracy reward. For relation classification, we employ a binary reward—assigning positive feedback for correct predictions and negative feedback for incorrect ones. In this subsection, we discuss the reward design for the \textbf{RC} task; details for the \textbf{TE} task are provided in the Appendix.

\paragraph{Format Reward:} The structured prompt template for RC is illustrated in Figure~\ref{box:RC_template}. We use regular expression extraction to enforce a standardized response format. The model is permitted to reason freely, without restrictions on the number or placement of \texttt{<think></think>} tags. Correctness is determined solely based on the content within the last pair of \texttt{<answer></answer>} tags. Furthermore, we require the answer to be in the form $y(e_1, e_2)$ or $y(e_2, e_1)$, where $y \in \mathcal{Y}$ denotes one of the predefined relation types. The format reward score $r_{\text{format}}$ is computed as:
\begin{equation*}
    r_{\text{format}} =
\begin{cases}
1, & \text{if format is correct} \\
-3, & \text{if format is incorrect}
\end{cases}
\end{equation*}

\paragraph{Metric Reward:} If the response passes the format evaluation, we assign an additional reward based on its RE accuracy. Specifically, we employ a rule-based scheme in which the model receives a positive reward for correct classifications and a negative reward for incorrect ones:
\[
r_{metric} =
\begin{cases}
    2, & \text{if } y_{\text{true}} = y_{\text{pred}} \\
    -1.5, & \text{otherwise}
\end{cases}
\]
This reward design is consistent with standard practices in rule-based reinforcement learning, incentivizing accurate predictions while penalizing errors.




\paragraph{Final Reward:}
The final reward \( r \) combines a \textit{format reward} (\( r_{\text{format}} \)) and a \textit{task-specific metric reward} (\( r_{\text{metric}} \)), and is defined as follows:
\[
r =
\begin{cases}
    r_{\text{format}} , & \text{if } r_{\text{format}} \neq 1 \\
    r_{\text{format}} + r_{\text{metric}}, & \text{if } r_{\text{format}} = 1
\end{cases}
\]
We exclude metric-based evaluation for outputs that do not conform to the required format, thereby enforcing strict format adherence prior to assessing task-specific performance.



\section{Experiment on Entity Classification}
\subsection{Dataset}
In this paper, we consider two relation classification datasets. The first is the public SemEval-2010 Task 8 dataset (Sem2010) \citep{hendrickx2019semeval}, a widely used benchmark for multi-class relation classification. The second is a human-annotated proprietary corpus constructed for the Mental Disease Knowledge Graph (MDKG). Key statistics of both datasets are summarized in Table \ref{Table:datasets}.

\begin{table}[h!]
    \centering
    \small 
    \renewcommand{\arraystretch}{1.15}
    \begin{tabular}{@{}l p{2.1cm} p{2.4cm}@{}}
        \toprule
        & \textbf{Sem2010} & \textbf{MDKG} \\
        \midrule
        \textbf{Relations} & 
        Component-Whole, Instrument-Agency, etc. &
        Hyponym of, Located in, Risk factor of, Treatment for, etc. \\
        \textbf{\# of classes} & 17 & 17 \\
        \textbf{Train/Test} & 8,353 / 500 & 10,033 / 500  \\
        \bottomrule
    \end{tabular}
    \caption{Relation types and train/test splits for the SemEval-2010 and MDKG datasets. The number of classes accounts for relation directionality. The complete table is provided in Appendix~\ref{sec:keystat}.}
    \label{Table:datasets}
\end{table}

Each dataset consists of sentence–relation pairs, along with an annotation guide defining each relation type. Table~\ref{table:MDKG-pair} presents an example sentence–relation pair and the corresponding relation definition.

\begin{table}[h!]
    \centering
    \small
    \renewcommand{\arraystretch}{1.15}
    \begin{tabularx}{\columnwidth}{@{}lX@{}}
        \toprule
        \textbf{Sentence:} & The \texttt{<e1>}hypothalamus\texttt{</e1>} may as a key brain region involve in the \texttt{<e2>}inflammatory related depressive-like behaviors\texttt{</e2>}. \\
        \midrule
        \textbf{Relation:} & hyponym-of (e1, e2) \\
        \midrule
        \makecell[l]{\textbf{Definition of}\\\textbf{hyponym-of:}} & 
        \makecell[l]{
        This relation can indicate a hierarchical \\ link, with X being a subordinate or specific \\ instance  of Y \ldots\\
        "X is hyponym of Y" has following types: \\
        (a) Direct Categorization: \ldots \\
        (b) Appositive Formulation: \ldots \\
        (c) Indicators of Inclusion or Example:  \ldots \\
        (d) Alternative Naming:  \ldots \\
        } \\
        \bottomrule
    \end{tabularx}
    \caption{An example sentence–gold standard pair from the MDKG dataset, along with the definition of the corresponding relation from the annotation guide, is shown below. Full relation definitions for both datasets are provided in Appendix~\ref{sec:reldef}.}
    \label{table:MDKG-pair}
\end{table}

\subsection{Main results}

\begin{table*}[h!]
    \centering
    \begin{tabular}{l cc cc}
        \toprule
        \multirow{2}{*}{\bf Model} 
        & \multicolumn{2}{c}{\bf MDKG (Private)} 
        & \multicolumn{2}{c}{\bf Sem-2010 (Open)} \\
        & Avg@4 & Pass@4 & Avg@4 & Pass@4 \\
        \midrule
        \multicolumn{5}{c}{\textbf{\textit{Proprietary Models}}} \\
        Claude 3.5 Sonnet & 71.1 $\pm$ 0.1   & 99.2    & 81.3 $\pm$ 0.1    & 100    \\
        GPT-4o  & 65.9 $\pm$ 0.1   & 99.4    & 79.8 $\pm$ 0.0    & 100    \\
        GPT-4.1-mini & 70.2 $\pm$ 0.1    & 99.6    & 81.0 $\pm$ 0.2    & 99.8    \\
        \midrule
        \multicolumn{5}{c}{\textbf{\textit{Open Source Models}}} \\
        Qwen-2.5-7B-Instruct                &  35.2 $\pm$ 1.2 & 48.4 & 38.6 $\pm$ 1.2 & 59.8 \\
        Llama-3-8B-Instruct                 &  22.7 $\pm$ 0.0 & 53.8 & 25.6 $\pm$ 0.1 & 56.2 \\
        Qwen-2.5-14B-Instruct               & 41.3 $\pm$ 0.8 & 56.2 & 54.6 $\pm$ 1.4 & 74.2 \\
        Qwen-2.5-32B-Instruct               & 49.2 $\pm$ 2.1 & 64.8 & 66.4 $\pm$ 1.2 & 83.8 \\
        Qwen-2.5-72B-Instruct               & 55.2 $\pm$ 1.5 & 73.6 & 70.2 $\pm$ 0.9 & 85.4 \\
        \midrule
        \multicolumn{5}{c}{\textbf{\textit{RL from Instruct Models}}} \\
        Qwen-2.5-7B-Instruct                &  & &  &  \\
        $\drsh$ R1-RE-7B (MDKG)             &  {88.1 $\pm$ 0.3} & {90.8} & \textbf{66.5 $\pm$ 0.2}  & \textbf{84.2} \\
        $\drsh$ R1-RE-7B (Sem)              & \textbf{65.8 $\pm$ 0.1} & \textbf{76.6} & {91.8 $\pm$ 0.0} & {100} \\
        \midrule
        \multicolumn{5}{c}{\textbf{\textit{SFT from Instruct Models}}} \\
        Qwen-2.5-7B-Instruct  & & & & \\
        $\drsh$ SFT-7B (MDKG) & {90.4$\pm$ 0.2} & {99} & \textbf{51.3$\pm$ 0.4} & \textbf{69.0} \\
        $\drsh$ SFT-7B (Sem)               & \textbf{49.8$\pm$ 0.1} & \textbf{65.3} & {92.4 $\pm$ 0.0} & {100} \\
        \bottomrule
    \end{tabular}
    \caption{Zero-shot relation classification accuracy of different models on the MDKG and Sem-2010 datasets. \textbf{R1-RE-7B (MDKG)} denotes the model trained on the MDKG dataset; the same naming convention applies to the SFT models. \textbf{Avg@4} indicates the average Pass@1 accuracy across 4 samples. For fair comparison, all the models use the Template in \ref{box:RC_template}. Out-of-domain accuracies are highlighted in \textbf{bold}.}
    \label{tab:RCmain}
\end{table*}

The main results are presented in Table~\ref{tab:RCmain}. Here, R1-RE-7B refers to our proposed method, with all evaluations conducted in a zero-shot setting using the prompt shown in Figure~\ref{box:RC_template}. All results are reported in terms of \textbf{Avg@4} accuracy, which denotes the average Pass@1 accuracy across four samples. The key observations are as follows:

\begin{itemize}[leftmargin=*]
\item The reinforcement learning process significantly enhances the relation extraction capabilities of the base model. R1-RE-7B substantially outperforms its backbone (Qwen-2.5-7B-Instruct) in both in-domain ($+52.9, 53.2$) and out-of-domain ($+27.9, 30.6$) accuracy.
\item Compared to its SFT counterpart, R1-RE demonstrates much stronger out-of-domain generalization, surpassing it by $+16.0$ and $+15.2$ points.
\item Both proprietary and open-source models achieve substantially higher accuracy on the Sem-2010 (public) dataset than on the MDKG (private) dataset, suggesting that the public benchmark may suffer from data leakage.
\item R1-RE-7B demonstrates OOD performance on the private MDKG dataset that is comparable to state-of-the-art proprietary models such as GPT-4o and GPT-4.1-mini. The relatively lower performance observed on the Sem-2010 dataset is likely attributable to data leakage, which artificially inflates accuracy of proprietary models.
\end{itemize}

\subsection{How R1-RE boosts RE Performance?}
\begin{figure*}[h!]
    \centering
    \includegraphics[width=0.9\textwidth]{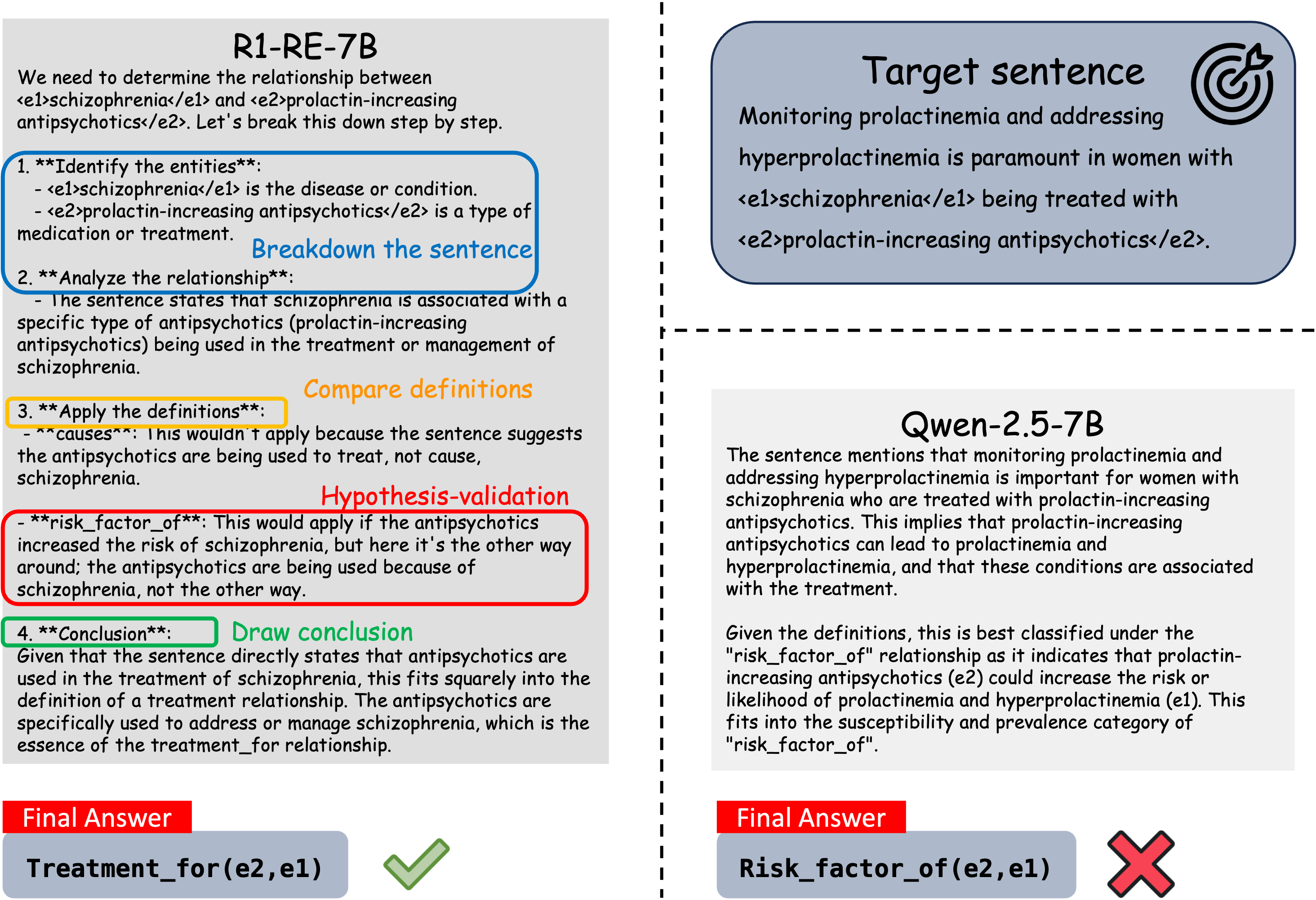}
    \caption{Case study comparing the chain-of-thought (COT) reasoning of \textbf{R1-RE-7B} and Qwen2.5-7B-Instruct. Due to space constraints, some COT outputs are omitted; the complete COT reasoning process for \textbf{R1-RE-7B} is provided in Appendix \ref{sec:COT}.}
    \label{fig:COT}
\end{figure*}

In this section, we further investigate a key question: \textit{How R1-RE improves RE performance?} To answer this question, we visualize the training dynamics of R1-RE and conduct a human analysis of its outputs. Figures~\ref{fig:training_summary} track response length, training reward, and both in-domain and out-of-domain (OOD) accuracy over the course of training, while Figure~\ref{fig:COT} compares outputs from the Qwen-2.5-7B-Instruct and R1-RE. We observe two key findings:

\begin{figure*}[h!]
    \centering
    \begin{subfigure}[t]{0.48\linewidth}
        \centering
        \includegraphics[width=\linewidth]{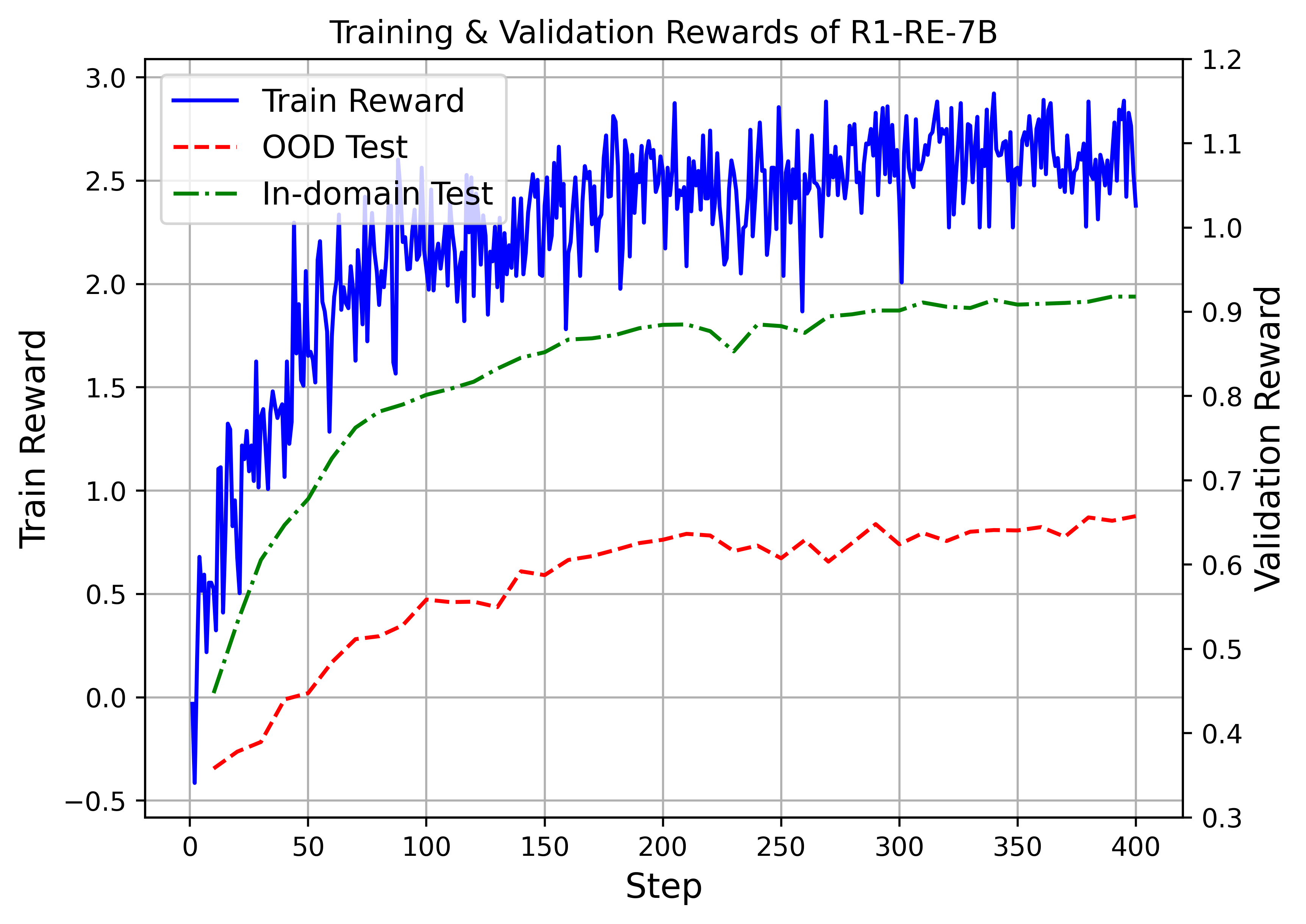}
        \caption{}
    \end{subfigure}
    \hfill
    \begin{subfigure}[t]{0.48\linewidth}
        \centering
        \includegraphics[width=\linewidth]{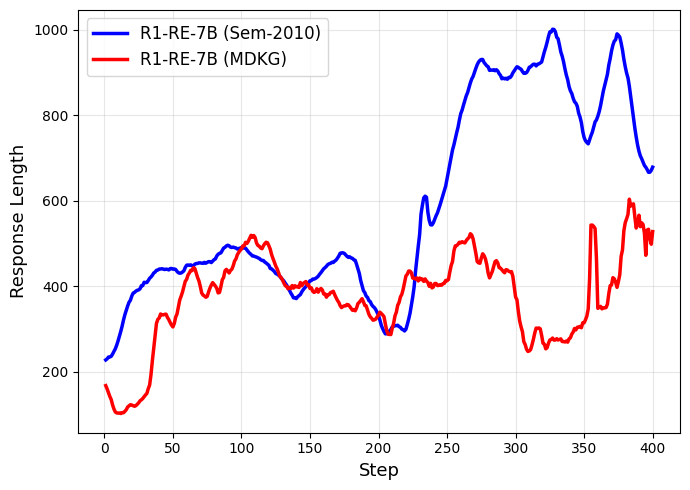}
        \caption{}
    \end{subfigure}
    \caption{(a) Training dynamics of R1-RE (Sem-2010), with the left y-axis representing the training reward and the right y-axis showing both in-domain and out-of-domain test accuracy. (b) Response length of R1-RE-7B models during training.}
    \label{fig:training_summary}
\end{figure*}
\begin{itemize}[leftmargin=*, itemsep=0pt, parsep=0pt, topsep=0pt, partopsep=0pt]
\item \textbf{Key Finding 1: Performance is correlated with both training rewards and response length.} Training rewards, response length, in-domain accuracy, and out-of-domain (OOD) accuracy all increase simultaneously throughout training. Specifically, compared with around 200 tokens of output at the beginning of the training process, the response length increases to $500$/$1000$  tokens on two datasets, indicating the emergence of long COT and aligns well with the phenomena in the existing literature \citep{xie2025logic}.
\item \textbf{Key Finding 2: R1-RE elicits genuinely human-like annotation behavior.} Unlike the baseline’s terse, pattern-matching answers, R1-RE-7B first identifies the entities in the context, then systematically compares each candidate relation with the definitions in the annotation guide, following a hypothesis–validation procedure; it finally draws a conclusion and outputs the answer. This step-by-step reasoning process closely mirrors the human annotation workflow depicted in Figure~\ref{fig:human}(b). This also indicates that the increased length shown in Figure \ref{fig:training_summary} is indeed a result of learning good annotation paradigms instead of simple overthinking~\citep{sui2025stop}. Notably, this reasoning patter emerges naturally during the RL training process, without the need for explicit distillation or supervised fine-tuning \citep{zheng2025learning}.
\end{itemize}

\subsection{Further Analysis}
\begin{itemize}[leftmargin=*, itemsep=0pt, parsep=0pt, topsep=0pt, partopsep=0pt]
    \item \textbf{The training process of R1-RE  preserves the model’s performance on other tasks.}
    Although {R1-RE} has acquired strong {RC} skills through reinforcement learning, its performance on other tasks remains unclear. Specifically, we are interested in whether the RC training process comes at a cost, such as reduced generalization or capability in unrelated domains. As highlighted in recent studies~\citep{zhao2023survey, kotha2023understanding, dai2025breach}, when LLMs are fine-tuned to excel at a particular skill, they may struggle to maintain performance in other areas. To investigate this trade-off, we benchmark \textbf{R1-RE} on three widely used datasets that evaluate distinct aspects of LLM ability: MATH-500 (mathematical reasoning)~~\citep{hendrycks2021measuring}, IFEval (instruction following)~~\citep{zhou2023instruction}, and GPQA (factual knowledge)~\citep{rein2024gpqa}.

    The results in Table~\ref{tab:generation} reveal a surprising yet reasonable trend. The RL process not only failed to suppress but actually improved the performance of {R1-RE} across all three benchmarks. In contrast, supervised fine-tuning (SFT) led to a noticeable decline in generalization performance. This finding is consistent with recent studies indicating that SFT tends to promote memorization of training data, thereby impairing out-of-domain generalization, whereas RL-based methods can facilitate better skill transfer and generalization across diverse tasks~\citep{chu2025sft}. 
    \begin{table}[htbp]
    \centering
    \resizebox{0.9\columnwidth}{!}{
    \begin{tabular}{lcc|c}
        \toprule
        \textbf{Model} & \makecell{\textbf{MATH}\\\textbf{500}} & \makecell{\textbf{IFEval}} & \makecell{\textbf{GPQA}} \\
        \midrule
        Qwen-2.5-7B & 73.6 & 72.3 & 29.2 \\
        R1-RE-7B & 75.4 & 73.0 & 29.6 \\
        SFT & 69.8 & 71.9 & 27.2 \\
        \bottomrule
    \end{tabular}
    }
    \caption{Performance of Qwen-2.5-7B-Instruct, \textbf{R1-RE}, and SFT models on three benchmarks. The evaluation is conducted with the default setting of the Evalscope framework.}
    \label{tab:generation}
    \end{table}

    \item \textbf{Incorporating additional training data can further improve the OOD performance of R1-RE.}
    Given the strong out-of-domain generalization exhibited by \textbf{R1-RE} on RC tasks, a natural question arises: how can we further improve its performance to match or even surpass that of proprietary models?

    Leveraging the broad availability of public RC datasets, we investigate the effect of incorporating additional data into training. Specifically, we include the SemEval-2018 Task 7 dataset~\citep{buscaldi2017semeval} (denoted as Sem-2018), training R1-RE on the combined dataset and re-evaluating its out-of-domain (OOD) accuracy. Remarkably, the inclusion of Sem-2018 leads to a substantial 4\% improvement in OOD performance on both the MDKG and Sem-2010 datasets. This result demonstrates the significant potential of our approach to further enhance generalization by integrating diverse and complementary datasets.

\end{itemize}

\begin{table}[h!]
    \centering
    \resizebox{0.9\columnwidth}{!}{
    \begin{tabular}{lcc}
        \toprule
        \textbf{Model} & \makecell{\textbf{Avg@4}\\\textbf{(MDKG)}} & \makecell{\textbf{Avg@4}\\\textbf{(Sem-2010)}} \\
        \midrule
        Qwen-2.5-7B-Instruct & 35.2 & 38.6   \\
        $\drsh$ R1-RE (MDKG) & \textbackslash{} & 66.5  \\
        $\oplus$ Sem-2018 & \textbackslash{} & \textbf{70.8}  \\[0.5ex]
        $\drsh$ R1-RE (Sem-2010) & 65.8 & \textbackslash{}  \\
        $\oplus$ Sem-2018 & \textbf{69.7} & \textbackslash{}  \\
        \bottomrule
    \end{tabular}
    }
    \caption{Out-of-domain (OOD) performance of R1-RE-7B after incorporating the additional Sem-2018 dataset.}
    \label{tab:model-comparison}
\end{table}

\section{Related works}
\label{sec:related}
\subsection{LLM reasoning}

LLM reasoning has garnered significant attention, as it demonstrates the potential of LLMs to generalize to complex real-world problems through human-like reasoning. Early efforts primarily focused on prompting methods, such as “chain of thought”\citep{wei2022chain} and “tree of thought”\citep{yao2023tree}, to elicit step-by-step reasoning. More recently, research has shifted toward explicitly training LLMs to master reasoning processes. Initial approaches often relied on reward models—such as outcome-based (ORM) or process-based (PRM) reward models~\citep{uesato2022solving}—but these methods can suffer from issues like reward hacking.

Recent advances, such as DeepSeek-R1~\citep{guo2025deepseek} and Vision-R1~\citep{huang2025vision}, have demonstrated that applying Reinforcement Learning with Verifiable Reward (RLVR) can effectively guide LLMs toward self-emergent reasoning without requiring trained reward functions or step-level human annotation. RLVR has been explored across a variety of domains, including logic games~\citep{xie2025logic}, search~\citep{jin2025search}, and machine translation~\citep{feng2025mt}. However, its application to knowledge extraction or relation extraction tasks remains underexplored.

\subsection{Relation extraction}
Early approaches to relation extraction predominantly adopt a supervised classification framework. Pipeline-based methods follow a two-step procedure: first applying named entity recognition (NER), then performing relation classification on the identified entity pairs~\citep{cai2016bidirectional}. Alternatively, span-based methods frame RE as a token-level classification task~\citep{eberts2020span}. With the advent of LLMs, recent work has begun to leverage their in-context learning capabilities through few-shot learning~\citep{borchert2023core, xu2023unleash}. Other approaches seek to enhance performance using retrieval-augmented generation (RAG) methods~\citep{wan2023gpt}. In addition, several studies aim to further improve accuracy via supervised fine-tuning~\citep{wadhwa2023revisiting, ettaleb2025contribution, shi2024cre}.

\section{Conclusion}
In this work, we revisited the relation extraction task by reframing it as a reasoning process grounded in annotation guidelines. We proposed \textbf{R1-RE}, a framework that employs RLVR to strengthen the reasoning abilities of LLMs for RE tasks. Experiments on both public and private datasets demonstrate that R1-RE achieves substantial improvements in out-of-domain performance, highlighting promising directions for enhancing the adaptability of LLMs in relation extraction.

\section*{Limitations}
In this work, we primarily focus on relation classification (RC) tasks and leave the exploration of more complex triplet extraction (TE) tasks for future research. Additionally, our experiments are limited to 7B-parameter models due to computational constraints. Evaluating R1-RE with larger models represents an important direction for future work.



\bibliography{custom}

\appendix

\section{Additional Materials on RC task}
\subsection{Annotation Guide}
\label{sec:reldef}
The annotation guide for the MDKG dataset is provided in Table~\ref{tab:MDKGguide}, while the guidelines for the SemEval datasets are available on their official website.

\begin{table*}[ht]
\small 
\textbf{risk-factor-of:} 

The relation descried in S entails that the presence of X hightens the risk or possibility of Y.
"X is risk-factor-of Y" has following types:

(a) Direct Risk Factor: The statement "X is a risk factor for Y" indicates that X directly increases the risk of Y occurring.

(b) Cause and Effect: Statements like "X is a leading cause for Y" or "X is caused/induced by Y" emphasize X's causative role in leading to Y. Relevant keywords include "contribute to" and "result in".

(c) Susceptibility and Prevalence: Phrases like "X is prone to developing Y", "X is more prevalent in Y", "X predisposes to Y", "X is at high risk for Y", "higher likelihood of X", "increased odds of X" suggest that X increases the likelihood of Y occurring. Additionally, statements like "X (a specific characteristic group) has an increased Y" fall under this category.

(d) Predictive Relation: The phrase "The happening of X is a predictor of Y" indicates that the occurrence of X can be used to predict the likelihood of Y happening in the future.

(e) Adverse Effects: X is a treatment or intervention and Y is an adverse effect or complication of that treatment.

\textbf{help-diagnose:}
This relation type encompasses methods used for diagnosing diseases as well as for observing specific signs, genes, or health factors, and may include diagnostic markers.
"X help-diagnose Y" has following types:

(a) Methods of Measurement and Detection: Tag sentences like "Use X to estimate/measure/detect/evaluate/assess Y". Here, X is used as a tool or method to quantify, identify, or assess Y. Other important keywords include 'based on', 'according to', 'diagnosed by', 'classify', 'screening', 'score'. 

(b) Diagnostic Criteria: Phrases such as "meeting X for Y" suggest that X acts as a standard or criterion for diagnosing or defining Y.

(c) Aiding in Identification and Differentiation: Phrases indicating that X can 'help identify', 'distinguish', 'differentiate', 'discriminate', or 'classify' B (a disease), pointing to X's usefulness in differentiating Y from other conditions or in making a definitive diagnosis of Y.

\textbf{characteristic-of:}

This relations describe the symptoms, clinical manifestations, or distinct features of a disease. 
"X is characteristic-of Y" has following types:

(a) Characteristic Identification: Sentences where X is described with phrases like 'characterized by', 'symptom of', 'clinical expression', 'hallmark', signifying Y as a characteristic or symptom of X. For instance, "X is characterized by Y". 

(b) Manifestation Descriptions: Sentences where X(disease) is noted to 'present', 'show', 'exhibit', 'report', 'indicate', 'demonstrate', or 'have' a symptom, namely Y. For example, "X presents/shows/exhibits Y".

(c) Marker Identification: Sentences where X is identified by Y, or Y is mentioned as a marker or biomarker for X, or X is more sensitive to Y. For instance, "X is identified by Y" or "Y is a marker/biomarker for X".

(d) Accompaniment Patterns: Sentences where X is usually accompanied by Y. For example, "X with/accompanied by Y".

(e) Quantitative Changes: Sentences indicating that Y is higher, decreased, or increased in X (disease). For instance, "Y is higher/decreased/increased in X".

(f) Observation in Disease: Sentences where Y was observed, found, or occurs in X. For example, "Y was observed/found/occur in X".

\textbf{associated-with:}

This relation describes an association or connection between X and Y, where changes in X may correspond with changes Y or changes in Y correspond with X. The associated with relation doesn't have direction such that associated-with(e1,e2) and associated-with(e2,e1) are equivalent. It not necessarily imply a cause-and-effect relation.

\textbf{treatment-for:}

This relation type highlights the link between X(treatments or interventions) and their impact on Y(a specific disease or condition). It includes a range of treatment methods like medication, psychotherapy, lifestyle modifications, surgical procedures, and others.
"X is treatment-for Y" has following types:

1. Application in Treatment: Phrases like "Use X to treat Y", "X is a treatment/intervention for Y" or "Y treated/medicated with X" indicate the use of X in the treatment of Y.

2. Signs of Improvement: Phrases where X 'improves', 'alleviates', 'suppresses', 'shows efficacy in', 'benefits', 'prevents', or 'reverses' Y, suggesting a beneficial impact of X on Y.

\textbf{hyponym-of:}

This relation can indicate a hierarchical link, with X being a subordinate or specific instance of Y, or a constitutive link, showing that X is a component or part of Y.
"X is hyponym-of Y" has following types:

(a) Direct Categorization: Sentences where "X is a Y" are used, signifying that X belongs to the category or class of Y.

(b) Appositive Formulation: Sentences in the form of "X, a Y", clearly identifying X as a type or instance of Y.

(c) Indicators of Inclusion or Example: Phrases using 'include', 'involve', 'such as', 'like', 'e.g.', to demonstrate that X is an example or part of the broader category Y.

(d) Alternative Naming: The use of "also called" to connect X with Y, indicating that X is another name or term for Y.

(e) Usage of Colon for Definition: Sentences that use a colon (":") to lead into a description, often employed to list or define hyponyms or specific instances under a broader category.

\textbf{abbreviation-for:}

This relation type refers to the use of X(abbreviations or acronyms) to represent Y(longer terms or phrases), especially in medical or technical fields. For example, "AIDS" is an abbreviation for "Acquired Immune Deficiency Syndrome", and "MRI" stands for "Magnetic Resonance Imaging".

\textbf{occurs-in:}

The occurs in relation applies to scenarios where X(an event, condition, or phenomenon) is observed in Y(groups or settings). Example: "Condition X occurs in populations with attributes Y".

\textbf{located-in:}
"X is located-in Y" occurs in following three scenarios:

(a) Gene or Region Specificity: This relation applies when discussing the location of genes or specific regions. For instance, "Gene/region X is located in gene/region Y".

(b) Disease, Symptom, or Physiological Activity Location: highlights the physical or anatomical location of medical conditions or biological processes. For example, "Disease/symptom/physiological activity X occurs in region Y".

(c) Localization Indicators for Signs: This relation also applies to sentences that indicate where signs of a particular condition (like a disease, gene, region, etc.) are found. For example, "X of Y (region/gene/…)".
\caption{The Annotation guide of the MDKG dataset.}
\label{tab:MDKGguide}
\end{table*}

\subsection{Complete reasoning output from \textbf{R1-RE-7B} and Qwen-2.5-7B-Instruct}
\label{sec:COT}
The COT reasoning outputs for an example prompt from \textbf{R1-RE-7B} and Qwen-2.5-7B-Instruct are shown in Table~\ref{tab:R1COT} and Table~\ref{tab:baseCOT}, respectively.

\begin{table*}[ht]
\small 
<think>

Let's analyze the given sentence: "Monitoring prolactinemia and addressing hyperprolactinemia is paramount in women with <e1>schizophrenia</e1> being treated with <e2>prolactin-increasing antipsychotics</e2>."

We need to determine the relation between <e1>schizophrenia</e1> and <e2>prolactin-increasing antipsychotics</e2>. Let's break this down step by step.                   

1. **Identify the entities**:

   - <e1>schizophrenia</e1> is the disease or condition.
   
   - <e2>prolactin-increasing antipsychotics</e2> is a type of medication or treatment.                                                                                      

2. **Analyze the relation**:

   - The sentence states that schizophrenia is associated with a specific type of antipsychotics (prolactin-increasing antipsychotics) being used in the treatment or management of schizophrenia.
   
   - We need to determine if this relation is that schizophrenia leads to or is treated by these antipsychotics, or if it's some other relation.                     

3. **Apply the definitions**:

   - **treatment-for**: This relation would apply if the antipsychotics are used to treat or manage schizophrenia. The sentence mentions that these antipsychotics are used specifically in the context of women with schizophrenia, implying a therapeutic use.
   
   - **risk-factor-of**: This would apply if the antipsychotics increased the risk of schizophrenia, but here it's the other way around; the antipsychotics are being used because of schizophrenia, not the other way.
   
   - **associated-with**: While it's possible that there could be an association, the sentence is more about the use of these antipsychotics as a therapeutic intervention rather than just an association.
   
   - **hyponym-of**: This would apply if prolatcin-increasing antipsychotics were a subtype or specific example of a larger category, but here it seems to be a specific type of antipsychotic used for a particular condition.
   
   - **causes**: This wouldn't apply because the sentence suggests the antipsychotics are being used to treat, not cause, schizophrenia.                                     
   - **characteristic-of**: This would describe a characteristic of the condition, but here we are describing the treatment rather than a characteristic of schizophrenia.   

4. **Conclusion**:

   - The sentence clearly states that these antipsychotics are being used in the context of women with schizophrenia. This implies that the antipsychotics are a form of treatment specifically for schizophrenia. The key phrase is "being treated with," which indicates that the antipsychotics are used to manage or treat the condition of schizophrenia.

Given that the sentence directly states that antipsychotics are used in the treatment of schizophrenia, this fits squarely into the definition of a treatment relation. The antipsychotics are specifically used to address or manage schizophrenia, which is the essence of the treatment-for relation.                                           

Therefore, the relation here is that the antipsychotics are used to treat schizophrenia.                                                                     

</think>
<answer>treatment-for(e2,e1)</answer>
\caption{An example of chain-of-thought (COT) reasoning output from R1-RE-7B}
\label{tab:R1COT}
\end{table*}

\begin{table*}[ht]
\small 
<think>
The sentence describes an interplay between DISC1 (intrinsic signaling pathway) and GABA signaling (extrinsic signaling pathway) in the context of neurogenesis, which is affected by susceptibility genes. The susceptibility genes here are DISC1 and GABA. The relation is about how these pathways influence neurogenesis and their impact on neuronal development and risk for mental disorders.

The main focus is on how the presence of DISC1 and GABA signaling, which are considered susceptibility pathways, influence neurogenesis. Since susceptibility pathways are risk factors for certain conditions (in this case, potentially mental disorders), we can infer that DISC1 is a risk-factor-of neurogenesis in the context of developmental processes and experiences.

However, the sentence does not directly indicate that DISC1 is a treatment or method for neurogenesis. Therefore, the relation "risk-factor-of" seems most appropriate here, as it captures the role of DISC1 in influencing the risk or likelihood of neurogenesis being affected by susceptibility genes.                                            

The direction of the relation is from DISC1 to neurogenesis, indicating that the presence of DISC1 (as a susceptibility pathway) impacts neurogenesis.                   
</think>

<answer>
risk-factor-of(e1,e2)
</answer>

\caption{An example of chain-of-thought (COT) reasoning output from Qwen-2.5-7B-Instruct}
\label{tab:baseCOT}
\end{table*}

\subsection{Key statistics of Sem2010 and MDKG datasets}
\label{sec:keystat}
\begin{table*}[h]
    \centering
    \renewcommand{\arraystretch}{1.2} 
    \begin{tabular}{l p{6cm} p{5.5cm}}
        \toprule
        & \textbf{Sem2010} & \textbf{MDKG} \\
        \midrule
        \textbf{Relations} & 
        \makecell[l]{Component-Whole, Cause-Effect \\ Instrument-Agency , Content-Container \\ Message-Topic , Product-Producer \\ Entity-Origin , Member-Collection \\ Other (no direction)}
        & 
        \makecell[l]{Hyponym of , Located in \\ Characteristic of , Abbreviation for \\ Risk factor of , Occurs in \\ Treatment for , Help diagnose \\ Associated with (no direction) } \\
        \textbf{\# of classes} & 17 & 17 \\
        \textbf{Train/Test} & 8,353/ 500 & 10,033/ 500 \\
        \bottomrule
    \end{tabular}
    \caption{Relation types and sample sizes of Sem2010 and MDKG datasets. Number of classes consider the directions of relations. }
    \label{Table: datasets_full stat}
\end{table*}

\section{Additional Materials on TE task}
\label{appendix:TE}
\textbf{Triplet Extraction(TE)}:  
In this task, the goal is to extract all valid triplets ($e_{\text{sub}}$: $t_{\text{sub}}$, $y$, $e_{\text{obj}}$: $t_{\text{obj}}$) from the sentence $\mathcal{S}$, where $t_{\text{sub}}, t_{\text{obj}} \in \mathcal{E}$ denote the entity types of the subject and object, respectively. A sentence $\mathcal{S}$ may contain multiple triplets. For instance

\begin{tcolorbox}[colback=white, colframe=black!20!white, boxrule=0.4pt, arc=1mm, left=1mm, right=1mm, top=0.5mm, bottom=0.5mm]
\textbf{\texttt{Sentence}} {Olanzapine was also associated with more frequent reports of weight gain and significantly greater VA costs \ldots}\\
\textbf{\texttt{Label}} {[[Olanzapine:drug, risk-factor-of, weight gain:symptom]]}
\end{tcolorbox}

\subsection{Reward Design}
\textbf{Format Reward:} For the TE task, the answer should be a list of triplets, each in the format $(e_{\text{sub}}{:} t_{\text{sub}}, y,e_{\text{obj}}{:} t_{\text{obj}})$, where $y \in \mathcal{Y}$ is a valid relation and $t_{\text{sub}}, t_{\text{obj}} \in \mathcal{E}$ are the corresponding entity types.

\noindent\textbf{Metric Reward:} The reward design for the \textbf{TE} task is inherently more complex, as it incorporates aspects of the named entity recognition (NER) task, requiring an evaluation of the extracted entities against gold standards.

Prior to the emergence of large language models (LLMs), traditional approaches typically adopted a strict criterion: an entity was considered correct only if the predicted span exactly matched the ground truth \citep{eberts2020span}. However, such evaluation protocols can be overly rigid when applied to LLMs. As noted by \citet{wadhwa2023revisiting}, this strictness often fails to account for the more flexible and nuanced outputs of LLMs. To address this limitation, \citet{wadhwa2023revisiting} employed human judges to provide a more accurate and fair assessment of model predictions. In this work, we allow the extracted entity to deviate by one token, either in the front or back, from the gold standard. We propose that a more flexible rule can be adopted, such as using a LLM as the judge \citep{gu2024survey}.

We propose a two-stage, F1 score-based as follows:
\begin{equation*}
r_{\text{TE}} = w_{\text{ent}} \cdot \text{F1}_{\text{ent}} + w_{\text{tri}} \cdot \text{F1}_{\text{tri}}.
\end{equation*}
Specifically, $\text{F1}_{\text{ent}}$ denotes the F1 score computed over the extracted entities. To compute $\text{F1}_{\text{ent}}$, we extract all unique (entity, entity type) pairs from both the ground truth and the model’s output, and calculate the precision and recall. $\text{F1}_{\text{tri}}$ represents the F1 score over the extracted triplets. For $\text{F1}_{\text{tri}}$, we directly compare the predicted and gold-standard triplets.

We make two key observations regarding our reward design. First, the RL framework enables us to directly employ the $F_1$ score as the reward, thereby eliminating the training–evaluation gap that arises in conventional approaches which rely on surrogate loss functions \citep{eberts2020span, shen2021trigger}. Second, our reward comprises two levels: the entity-level reward acts as an intermediate signal, providing early-stage feedback when the model is not yet capable of extracting complete triplets. This approach is supported both in theory and practice, as intermediate rewards are known to accelerate learning and enhance exploration in RL settings~\citep{ng1999policy, zhai2022computational}. The triplet-level $F_1$ score, on the other hand, reflects the final objective of the task. To ensure that learning remains focused on the end goal, we assign higher weight to the triplet-level reward ($w_{\text{ent}} = 1$, $w_{\text{tri}} = 3$), thereby mitigating the risk of divergence from the main objective.

\section{Prompt Design}
\label{sec:TE_template}

\begin{tcolorbox}[
    colframe=pink!80!black,
    colback=pink!10!white,
    fontupper=\small,
    boxrule=0.4mm,
    left=2mm, right=2mm, top=0.5mm, bottom=0.5mm,
    arc=2mm,
    enhanced
]
The user gives a sentence. The Assistant need to extract triplet from the sentence. Only consider the following entity types:\\
\{{\color{red}Annotation guide - Entity}\}\\
Only consider the following relation types:\\
\{{\color{red}Annotation guide - Relation}\}\\
The assistant first thinks out load and then provides the user with the final answer, make sure the final answer is enclosed within <answer> </answer> and only appear once. i.e., reasoning process here <answer> answer here </answer>. The answer should be a list of triplets, with each triplet have form [object:type, relation, subject:type].\\
User: \{{\color{red}Sentence}\}\\
\end{tcolorbox}

\noindent\begin{minipage}{\linewidth}
\captionof{figure}{The prompt for TE tasks.}
\end{minipage}

\end{document}